\documentclass[letterpaper]{article} 
\usepackage{aaai25}  
\usepackage{times}  
\usepackage{helvet}  
\usepackage{courier}  
\usepackage[hyphens]{url}  
\usepackage{graphicx} 
\usepackage{natbib}  
\usepackage{caption} 
\usepackage{algorithm}
\usepackage{algorithmic}
\usepackage{amssymb}
\usepackage{amsmath}
\usepackage{booktabs}
\usepackage{multirow}
\usepackage{array}
\usepackage{newfloat}
\usepackage{listings}
\DeclareCaptionStyle{ruled}{labelfont=normalfont,labelsep=colon,strut=off} 
\floatstyle{ruled}
\newfloat{listing}{tb}{lst}{}
\floatname{listing}{Listing}
\title{TGBFormer: Transformer-GraphFormer Blender Network \\ for Video Object Detection}
\author{
	Qiang Qi\textsuperscript{}, Xiao Wang\textsuperscript{}\thanks{Corresponding Author.}
} 
\affiliations{
	\textsuperscript{}School of Data Science, Qingdao University of Science and Technology, Qiangdao, China\\
	
qiangq@qust.edu.cn, xiaowang@qust.edu.cn
}

\usepackage{bibentry}

\begin{document}
	
	\maketitle
	
	\begin{abstract}
		Video object detection has made significant progress in recent years thanks to convolutional neural networks (CNNs) and vision transformers (ViTs).
		Typically,
		CNNs  excel at capturing  local features 
		but struggle to model global representations. 
		Conversely, ViTs 
are adept at
		capturing long-range global features 
		but face challenges in representing local feature details.
		Off-the-shelf video object detection methods solely rely on  CNNs or ViTs to conduct feature aggregation, which 
		hampers
		their capability to simultaneously leverage  global and local  information,
		thereby resulting in limited detection performance.
		In this paper, we propose a  Transformer-GraphFormer Blender Network (TGBFormer) for video object detection, with three key technical improvements to  fully exploit the advantages of transformers and graph convolutional networks while compensating for their limitations.
		First, we develop a spatial-temporal transformer module to aggregate global contextual information, constituting global representations with long-range feature dependencies.
		Second, we introduce a spatial-temporal GraphFormer module that utilizes local spatial and temporal relationships to aggregate features, generating new local representations that are complementary to the transformer outputs.
		Third, we design a global-local feature blender module to adaptively couple transformer-based global representations and GraphFormer-based local representations.
		Extensive experiments demonstrate that our TGBFormer establishes new state-of-the-art results on the ImageNet VID dataset. Particularly, our TGBFormer achieves 86.5\% mAP while running at around 41.0 FPS 
		on a single Tesla  A100 GPU.
	\end{abstract}
	
	%
	
	\section{Introduction}
	
	Video object detection
	aims to predict the location boxes and category labels for each object in videos. 
	It 
	plays an important role in a broad range of  applications, such as  
	safe driving \cite{1,2}, security surveillance \cite{3,4} and activity understanding \cite{5,6}.
	In the past decades, image object detection has achieved immense progress and delivered significant improvement in performance. 
	Unfortunately, 
	these well-built image object detectors suffer from remarkable performance drop when applied to video data, due to the appearance deterioration situations arising from motion blur, partial occlusion and unusual poses. 
	\begin{figure}[!t]
		\centering
					\includegraphics[width=0.474\textwidth,height=0.322\textwidth]{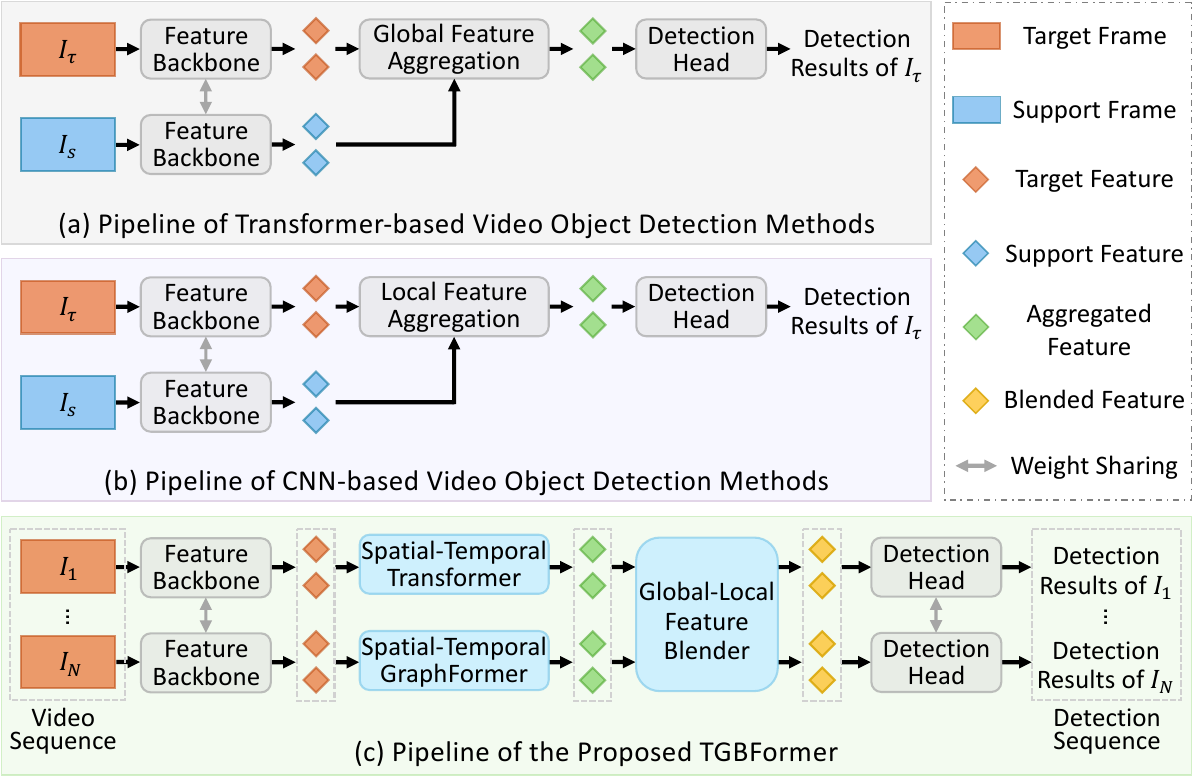} 
		\caption{Pipeline comparison of different  methods. 
		}
		\label{figure 1}
	\end{figure}

	Since videos come with extra temporal information compared to static images, it is intuitive  to  make full use of the temporal information to  alleviate the appearance deterioration situations.
	Based on this intuition, some off-the-shelf video object detection methods \cite{8,12,13,15}, whose pipeline is illustrated in Figure \ref{figure 1}(a), put efforts into leveraging  the long-range temporal dependencies captured by different types of vision transformers to aggregate features from a global perspective.
	Typically, 
	vision transformers are adept at capturing long-range global information but face challenges in representing short-range local information \cite{23,22,24}.
	Thus,
	these transformer-based video object detection methods do not fully consider short-range temporal dependencies  and thus lack the local perception capability for objects,
	which possibly leads to the  false detection problem.
	Some other existing video object detection methods \cite{9,17,16,21}, whose pipeline is shown in Figure \ref{figure 1}(b), 
	devote to
	exploiting the short-range temporal dependencies provided by different types of convolutional neural networks (CNNs) to aggregate features from a local perspective.
	However, CNNs are observed to excel at modeling short-range local information but struggle to capture long-range global information \cite{22,24}. 
	Thus,
	these CNN-based video object detection methods do not take  long-range temporal dependencies into consideration and thus lack the global perception capability for  objects,
	which may result in the missed detection problem.
	In contrast, the human visual system can simultaneously leverage  long-range and short-range temporal information to achieve more comprehensive perception capability for objects.
	Such a fact 
	motivates us to combine the advantages of transformers and CNNs to improve  the  video object detection task, which has not  been investigated in previous video object detection works.
	
	In this paper, we propose a Transformer-GraphFormer Blender Network (TGBFormer) for video object detection, 
	whose pipeline is illustrated in Figure \ref{figure 1}(c),
	including
	three key technical improvements to sufficiently combine the advantages of transformers and graph convolutional networks while compensating for their limitations.
	Although it is easy to think of combining the transformer and GraphFormer models, a simple ensemble of these two models is costly and yields marginal performance improvements. 
	To achieve an effective and efficient combination of them, there are three crucial problems that need to be addressed:  
	i) how to customize the transformer and GraphFormer to aggregate beneficial  information, constituting the global  and local representations of objects; 
	ii) how to couple the transformer global representations and GraphFormer local representations in a complementary and interactive paradigm,
	retaining the global modeling capability of transformers and local modeling capability of CNNs to the maximum extent;
	iii) how to improve inference/running  speed,
	striking a balance between detection accuracy and inference speed.
	To address the first problem, 
	we develop a spatial-temporal transformer module that exploits long-range spatial and temporal dependencies to aggregate beneficial features, constituting the global representations of objects.
	Moreover, 
	we present a spatial-temporal GraphFormer module that utilizes short-range spatial and temporal relationships to aggregate useful features, generating new local representations that are complementary to the transformer outputs.
	To handle the second problem, we design a global-local feature blender module to adaptively couple the transformer global representations and GraphFormer local representations, 
	retaining the global perception capability of transformers and local modeling capability of CNNs to the maximum extent.
	To solve the third problem,  contrary to the non-parallel frame-wise detection fashion that is commonly used in previous video object detection works,
	we  adopt a parallel  sequence-wise detection fashion in view of the advantage of  TGBFormer's comprehensive (including both global and local) object perception capability, which  simultaneously detects objects on all input frames and thus significantly improves inference speed.
	
	The above customized components are closely integrated into a uniform framework, allowing our TGBFormer to effectively combine the strengths of transformers and graph convolutional networks to produce more comprehensive appearance representations, which facilitates improving video object detection performance.
	To the best of our knowledge, our TGBFormer is the first effort that 
	exploits the complementarity of transformer global information and CNN local information  to tackle the  video object 
	detection task.
	
	We summarize the key contributions as follows:
	\begin{itemize}
		\item 
		We propose a novel Transformer-GraphFormer Blender Network (TGBFormer) for video object detection, which 
		combines the merits of 
		transformers and graph convolutional networks while compensating for their limitations,
		thereby establishing new state-of-the-art  results on the  ImageNet VID dataset.
		
		\item We introduce
		two collaborative modules, termed  spatial-temporal transformer and GraphFormer modules,
		that use global and local spatial-temporal dependencies to aggregate features respectively,
		enabling our TGBFormer  to harness the capability of transformers to  capture global information 
		while embracing the power of graph convolutional networks to model local information.

		\item  We design a  global-local feature blender module to fuse transformer-style global features and GraphFormer-style local features in a complementary  fashion, 
		making our TGBFormer preserve the global  and local  perception capabilities  for objects to the maximum extent.
		
	\end{itemize}

	\begin{figure*}[t]
		\begin{center}
			\includegraphics[width=1.0\textwidth]{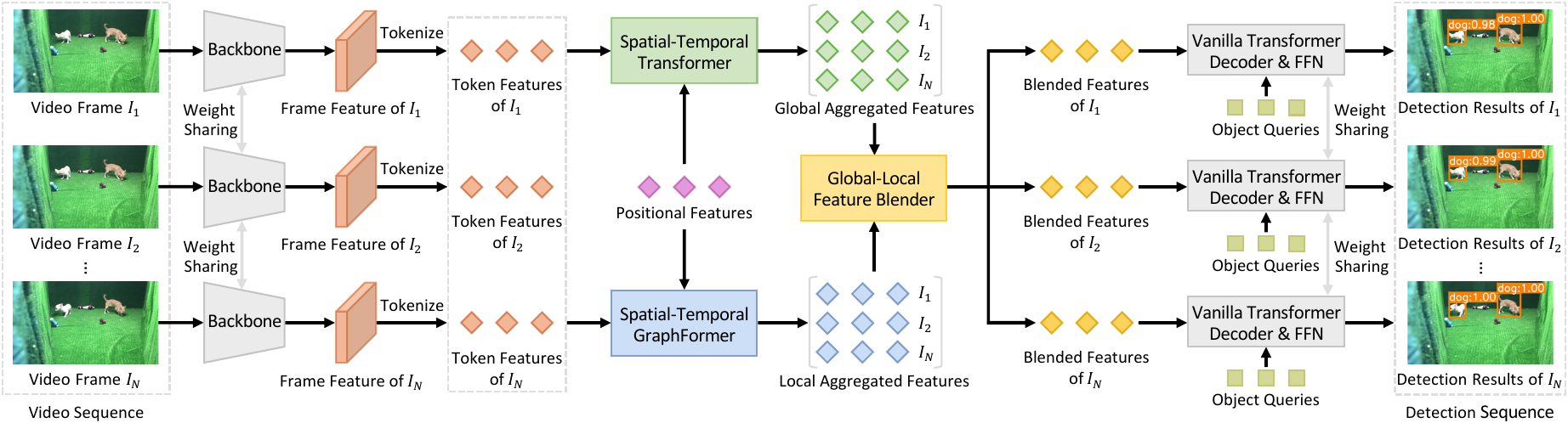}
		\end{center}
		\caption{%
			Framework of the proposed TGBFormer, 
			which builds upon the DETR \cite{26} model.
		}
		\label{figure 3}
	\end{figure*}

	\section{Related Work}
	
	Research on video object detection 
	has proceeded along two directions:
	CNN-based video object detection  methods and transformer-based video object detection methods.
	
	CNN-based video object detection methods \cite{9,17,16,21} usually put efforts into leveraging the temporal dependencies captured by different types of convolutional neural networks (CNNs) or operations to aggregate beneficial features.
	Typically,
	CNNs  excel at capturing short-range information but face challenges in modeling long-range  information \cite{22,23,24}.  
	Thus, 
	these CNN-based video object detection methods fail to fully exploit long-range temporal dependencies, which limits their global  perception capability for objects.

	Transformer-based video object detection methods \cite{8,13,14,15} generally put efforts into leveraging  the long-range temporal dependencies captured by different types of vision transformers to aggregate beneficial features.
	However, transformers are observed to excel at modeling long-range information but struggle to capture short-range information \cite{23,22,24}. 
	Thus,
	these transformer-based video object detection methods do not fully exploit short-range temporal dependencies, limiting their local perception capability for objects.
	
	Different from these methods that solely rely on CNNs or transformers to model temporal information,
	our TGBFormer customizes CNNs  and transformers in a unified video object detection framework, and combines the merits of both CNNs  and transformers to simultaneously explore short-range and long-range temporal information.
	
	\section{The Proposed Method}
	\label{section 3}
	\subsection{TGBFormer Framework}
	\label{section 3.1}
	
	Figure \ref{figure 3} illustrates the  framework of our TGBFormer, which mainly includes three elaborately-customized modules to sufficiently exploit the merits of transformers and graph convolutional networks while compensating for their limitations.
	Given an input video sequence, each video frame is first processed by a shared backbone network (e.g., ResNet-101) to generate the frame features, followed by a tokenization operation to produce the token features.
	Then, these token features together with the corresponding positional features (predicted by DETR \cite{26}) are fed into a spatial-temporal transformer module (STTM) to generate the global aggregated features.
	The carefully-customized STTM  utilizes  the spatial and temporal transformers to explore  long-range spatial-temporal dependencies and aggregate beneficial features, 
	enabling our TGBFormer to embrace the global  perception capability for objects. 
	Meanwhile, a spatial-temporal GraphFormer module (STGM) is developed to explore short-range spatial-temporal dependencies and aggregate useful features by using the spatial and temporal dynamic graph convolutional networks, 
	generating the local aggregated features that are complementary to the outputs of STTM
	and thus making our TGBFormer have  the local perception capability for objects. 
	After that, 
	a global-local feature blender module is designed to couple the global and local aggregated features in a complementary and interactive fashion, outputting the blended features. 
	Finally, 
	the blended features together with the object queries (generated by DETR) are fed into a vanilla transformer decoder and  feed forward network (FFN) to predict detection results.
	In view of the advantages of our elaborately-customized modules, 
	a parallel sequence-wise detection fashion is adopted in the framework, 
	allowing to simultaneously detect objects on all input  frames and thus endowing our TGBFormer with real-time inference speed.

	\subsection{Spatial-Temporal Transformer Module}
	\label{section 3.2}

\begin{figure}[!t]
		\centering
\includegraphics[width=0.474\textwidth,height=0.19\textwidth]{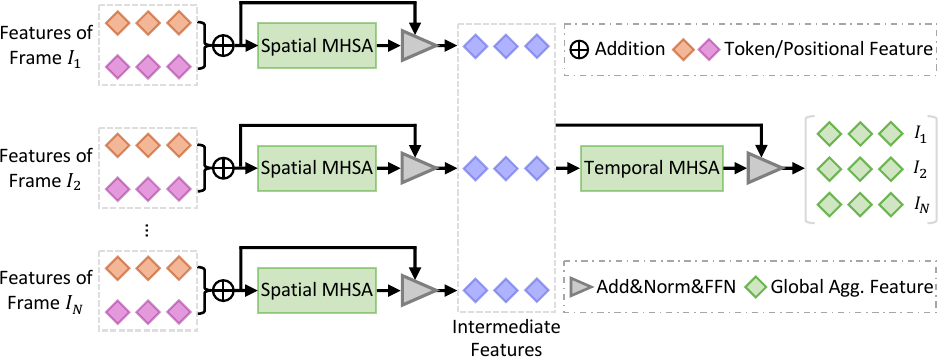} 	
		\caption{Illustration of the proposed STTM.}
		\label{figure 4}
	\end{figure}

	The illustration of the proposed STTM is presented in Figure \ref{figure 4}.  
	Given the frame feature $ f_{n} \in \mathbb{R}^{c\times h \times w }$  of the $n$-th frame extracted by the backbone network, in which $c$, $h$ and $w$ respectively denote the dimension, height and width of the frame feature,
	we first tokenize it  into non-overlapping $M\!$ $(=\!h \times w)$ token features with the dimension of $D$.
	Then, 
	the token features and the corresponding positional features (generated by DETR \cite{26}) of each frame are added together and  fed into a spatial multi-head self-attention block to effectively explore long-range spatial dependencies within a frame and perform  intra-frame feature aggregation from a global perspective.
	Building on the success of \cite{64},
	the calculation of the spatial multi-head self-attention (SpatMHSA) can be formulated
	as:
	\begin{equation}
		\text{SpatMHSA}(z_{q},x)  = 
		\sum_{t=1}^{T} \boldsymbol{W}_{\!t} \bigg[ \sum_{k=1}^{K} O_{tqk} \cdot \boldsymbol{W}_{\!t}^{'}  x_{k} \bigg],
		\label{equation 1}
	\end{equation}
	where 
	$q \in \Omega_{q}$ denotes the query element with the representation feature $z_{q} \in \mathbb{R}^{D}$, and 
	$k \in \Omega_{k}$ indicates the  key element with the representation feature  $x_{k} \in \mathbb{R}^{D}$. Here, $D$ denotes the feature dimension of $z_{q}$ and $x_{k}$.
	$\Omega_{q}$ and $\Omega_{k}$ represent the set of query and key elements, respectively.
	The case where $\Omega_{q}=\Omega_{k} $ is usually referred to as self-attention; 
	otherwise, it is referred to as  cross-attention.
	To disambiguate different spatial positions,
	the features $z_{q}$  and $x_{k}$ are the element-wise addition of the token and positional features
	in the implementation of our SpatMHSA.
	$T$ denotes the total number of attention heads, and $K$ represents the total number of key elements.
	$\boldsymbol{W}_{\!t} \in \mathbb{R}^{D \times D_{v}}$ and $ \boldsymbol{W}_{\!t}^{'} \in \mathbb{R}^{D_{v} \times D}$ denote the learnable projection weights, in which $D_{v}=D/T$.
	The notation of $\cdot$ indicates the scalar multiplication operation.
	$O_{tqk}$ represents the self-attention weights of the $k$-th sampling element in the $t$-th attention 
	head. 
	It is calculated by the scaled dot product between $z_{q}$ and $x_{k}$, and is normalized over all key elements:
	\begin{equation}
		O_{tqk}  \propto
		\text{exp} \bigg(\frac{z_{q}^{\mathsf{T}}  \boldsymbol{U}_{\!t}^{\mathsf{T}} \boldsymbol{V}_{\!\!t} x_{k}} {\sqrt{D_{v}}} \bigg),  ~~
		\sum_{k=1}^{K} O_{tqk} =1,
		\label{equation 2}
	\end{equation}
	where $\propto$  indicates the proportionality operation, and $(\cdot)^{\mathsf{T}}$ is the transpose operation.
	$\boldsymbol{U}_{\!t} \in \mathbb{R}^{D_{v} \times D}$ and $\boldsymbol{V}_{\!\!t} \in \mathbb{R}^{D_{v} \times D}$ are the learnable projection weights.
	Next, 
	the output features of SpatMHSA are passed through an addition, normalization and feed forward network to yield the intermediate features.
	
	After that, the intermediate features of each frame are fed into a temporal multi-head self-attention  block, 
	with the goal of effectively  excavating long-range temporal dependencies across frames and performing inter-frame feature aggregation from a global perspective.
	Particularly,
	the calculation of the temporal multi-head self-attention (TempMHSA) can be 
	represented by:
	\begin{equation}
		\begin{aligned}
			\text{TempMHSA}(z_{q}, \{x^{n}\}_{n=1}^{N}) 
			~~~~~~~~~~~~~~~~&  \\
			= \sum_{t=1}^{T} \boldsymbol{W}_{t}  \bigg[ \sum_{n=1}^{N}  \sum_{k=1}^{K} O_{tnqk} \cdot  \boldsymbol{W}_{\!t}^{'} x^{n}_{k} \bigg],
		\end{aligned}
		\label{equation 3}
	\end{equation}
	where $n$ indexes the input frame, and
	$O_{tnqk}$ denotes the self-attention weights of the $k$-th sampling element in the $t$-th attention head and the $n$-th frame.
	The  
	mathematical 
	calculation of $O_{tnqk}$ can be referred to as Equation (\ref{equation 2}), and 
	it can be correspondingly normalized by $\sum_{n=1}^{N} \sum_{k=1}^{K} O_{tnqk} =1$.
	The other symbols in Equation (\ref{equation 3}) have the same meanings as those in Equation (\ref{equation 1}).
	Finally,
	the output features of TempMHSA are sequentially fed into an addition and  normalization operation, followed by a  feed forward network  to generate the global aggregated features.

	\subsection{Spatial-Temporal GraphFormer Module}
	\label{section 3.3}

\begin{figure}[!t]
		\centering
				\includegraphics[width=0.474\textwidth, height=0.2676\textwidth]{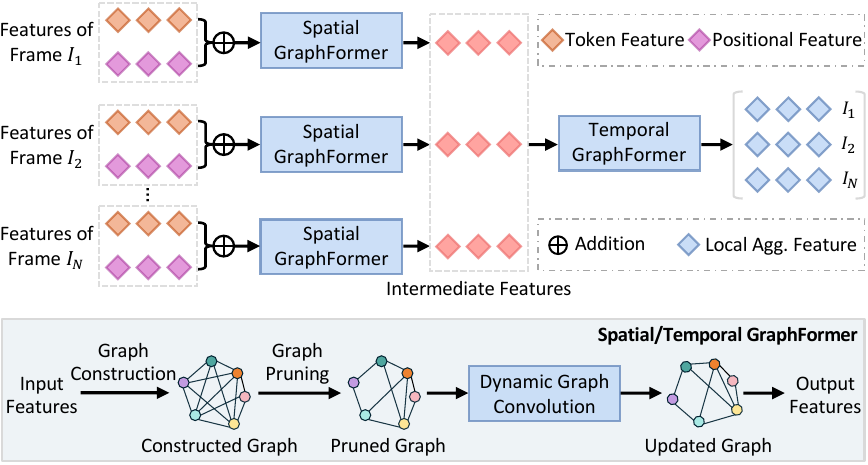} 
		\caption{Illustration of the proposed STGM.}
		\label{figure 5}
	\end{figure}

	The illustration of  the proposed STGM is depicted  in Figure \ref{figure 5}. 
	Specifically,
	we first tokenize the frame feature $ f_{n} \in \mathbb{R}^{c\times h \times w }$  of the $n$-th frame into non-overlapping $M$ 
	$ (=h \times w)$ 
	token features with the dimension of $D$.
	Then,
	the token features and the corresponding positional features of each frame are added together and  passed through a spatial GraphFormer block to explore short-range spatial dependencies within a frame and conduct intra-frame feature aggregation from a local perspective. 
	Specifically, the spatial GraphFormer first constructs a fully-connected undirected graph $\mathcal{G}(\boldsymbol{\mathcal{V}}, \boldsymbol{\mathcal{E}}, \boldsymbol{A}) $, in which
	$\boldsymbol{\mathcal{V}}$,
	$\boldsymbol{\mathcal{E}} $  and 
	$\boldsymbol{A} \in \mathbb{R}^{M \times M}$ respectively indicate the node set, edge set and adjacency matrix.
	In the construction of the graph,
	each input feature $r_{i} \in \mathbb{R}^{D}$ (i.e., the addition of each token and positional features) is assumed to be a graph node $v_{i} $,
	and each edge $e_{ij} $ is defined as the pairwise relationship between the nodes $v_{i}$ and $v_{j}$.
	Particularly, the edge $e_{ij}$ can be calculated as:
	\begin{equation}
		\!\!\!\!	e_{ij} = 
		\boldsymbol{\mathcal{F}}_{map}
		\big( \varPi \big [ Euc (r_{i},r_{j}), Cos (r_{i}, r_{j}), Sec (r_{i},r_{j}) \big] \big),\!\!\!
		\label{equation 4}
	\end{equation}
	where $	\boldsymbol{\mathcal{F}}_{map}(\cdot)$ denotes a mapping function consisting of two fully-connected layers, and $\varPi [\cdot,\cdot,\cdot]$ represents the concatenation operation.
	$Euc(\cdot,\cdot)$ means the standardized Euclidean distance, and 
	$Cos(\cdot,\cdot)$ denotes the cosine similarity. 
	$Sec(\cdot,\cdot)$ represents the semantic similarity, 
	and $Sec(r_{i},r_{j})$ can be calculated as $ r_{i}^{\mathsf{T}}r_{j}$.
	Based on the well-built edge $e_{ij}$,  the adjacency matrix $\boldsymbol{A} $ can be obtained through a softmax  function, in which $ \boldsymbol{A}_{ij} =\text{exp}(e_{ij}) / \sum\nolimits_{j=1}^{M} \! \text{exp}(e_{ij})$.
	
	Since the goal of our STGM is to 
	perform feature aggregation from the local perspective of graph convolutions,
	directly utilizing  a  fully-connected graph cannot well model the local relationships between node features,  in which case conducting graph convolutions may result in the aggregation of useless information (e.g.,  inter-class objects and backgrounds) and introduce much additional computational cost.
	To alleviate these problems,
	we present a graph pruning mechanism, 
	with the goal of sparsifying the topological structure of the graph to attend the useful local relationships between nodes by removing useless/weak connection edges.
	Different from previous methods that employ a single cut-off threshold in the normalized adjacency matrix to remove edges, which may result in indistinguishable feature representations due to the elimination of matrix elements below the threshold,
	our customized graph  pruning mechanism proposes to leverage  $S$ real-value thresholds denoted as $\varGamma=[\theta_{1}, \theta_{2}, \cdot\cdot\cdot, \theta_{S}] $, in which $\theta_{i}<\theta_{j}$ and $\theta_{i}, \theta_{j} \in [0,1]$ for $\forall i<j$.
	In light of this,
	we construct an adjacency tensor 
	$ \mathbb{A} \in  \mathbb{R}^{S\times M \times M}$  that consists of a set of adjacency matrices $ \{\mathbb{A}^{s}\}_{s=1}^{S}$, where $ \mathbb{A}^{s}\in \mathbb{R}^{M\times M}$.
	Particularly,
	we set $\mathbb{A}^{1} $ as the identity matrix $\boldsymbol{E} $,
	and for each $s \geqslant 2$, the mathematical calculation of $\mathbb{A}^{s} $ can be defined as:
	\begin{equation}
		\mathbb{A}^{s}_{ij} =
		\begin{cases} 
			\vspace{1ex}
			\boldsymbol{A}_{ij}  & \text{if }~\!    \theta_{s-1} \!\leqslant \boldsymbol{P}_{ij} \!< \theta_{s},  i \neq j  \\
			0 &  \text{Otherwise}
		\end{cases} ,
		\label{equation 5}
	\end{equation}
	where $\boldsymbol{P}_{ij} $ is an element in the probability matrix $\boldsymbol{P}$ and it can be calculated by $\boldsymbol{P}_{ij} = \lambda \cdot	\boldsymbol{A}_{ij}/d_{i}$, in which $\lambda$ (is set as 0.3) denotes a scalar weight and $d_{i}= \boldsymbol{\mathcal{D}}_{ii} =\sum_{j=1}^{M}	(\boldsymbol{A}+\boldsymbol{E})_{ij}$ is an element in the diagonal matrix $	\boldsymbol{\mathcal{D}} $.
	Next,
	building on the success of \cite{30}, we  softly select two adjacency matrices $\boldsymbol{Q_{1}}$ and $\boldsymbol{Q}_{2}$ from the adjacency tensor $\mathbb{A} $ via two $1\times 1 $ convolutions with non-negative weights  derived from the softmax function:
	\begin{equation}
		\!\!\!\!\!\boldsymbol{Q}_{1}\!  \!=\! \phi_{1}\! (\mathbb{A}, \text{softmax}(\boldsymbol{W}_{\!\phi_{1}}\!)), 
		\boldsymbol{Q}_{2}  \!=\! \phi_{2} (\mathbb{A}, \text{softmax}(\boldsymbol{W}_{\!\phi_{2}}\!)),\!\!\!\!\!
		\label{equation 6}
	\end{equation}
	where $\phi_{1}(\cdot,\cdot)$ and  $\phi_{2}(\cdot,\cdot)$ denote two different $1\times 1$ convolution layers.
	$\boldsymbol{W}_{\!\phi_{1}} \in \mathbb{R}^{1\times 1 \times S}$ and $ \boldsymbol{W}_{\!\phi_{2}}\in \mathbb{R}^{1\times 1 \times S}$ denote learnable  parameters.
	Subsequently,
	the final pruned adjacency matrix $	\boldsymbol{\bar{A}} \in  \mathbb{R}^{ M \times M}$ can be calculated as:
	\begin{equation}
		\boldsymbol{\bar{A}}  = \psi (\boldsymbol{Q}_{1}\boldsymbol{Q}_{2}+\boldsymbol{E}),
		\label{equation 7}
	\end{equation}
	where $\psi (\cdot)$ denotes the graph Laplacian normalization operation, 
	which can be represented by
	$ \psi (\boldsymbol{Y})=	\boldsymbol{\mathcal{D}}^{-\frac{1}{2}}  \boldsymbol{Y}	\boldsymbol{\mathcal{D}}^{-\frac{1}{2}}$.
	Empowered by Eqs. (\ref{equation 5})-(\ref{equation 7}), we can obtain the pruned graph.
	
	Based on the pruned graph, we extend the native graph convolutional network \cite{25} from the static setting to the dynamic setting, and customize a dynamic  graph convolution block with  residual connections  to  effectively aggregate beneficial information from local neighbors.
	Particularly,
	the customized dynamic 
	graph convolution block (DGCB) can be defined as:
	\begin{equation}
		\text{DGCB}(\boldsymbol{H})  = \text{DGCL}_{2}(\text{DGCL}_{1}(\boldsymbol{H}))+ \rho \cdot \boldsymbol{H},
		\label{equation 8}
	\end{equation}
	where $\boldsymbol{H} \in \mathbb{R}^{D \times M}$ is the node feature matrix of the pruned graph. 
	$\text{DGCL}_{1}(\cdot) $ and $\text{DGCL}_{2}(\cdot) $ denote  two sequential dynamic graph convolution layers, which improve the native graph convolution calculation in \cite{25}.
	In detail,
	different from the graph convolution in \cite{25} that uses a fixed  adjacency matrix for all layers,
	our customized dynamic graph convolution re-calculates the adjacency matrices  according to the newly learned node feature matrices at each layer, which facilitates capturing the dynamic graph structures during the graph convolution process and thus improves the effectiveness of the graph convolution. 
	$\rho$  denotes  the residual constant, which is empirically set as 0.5.
	As illustrated in Figure \ref{figure 5}, the outputs of the spatial GraphFormer  are denoted as  intermediate features.

	After that, the intermediate features of each  frame are passed through a temporal GraphFormer  block to  excavate short-range temporal dependencies across frames and perform inter-frame feature aggregation from a local perspective,
	generating the local aggregated features that are complementary to the outputs of STTM.
	The spatial GraphFormer and temporal GraphFormer  share the same workflow, but their 
	difference lies in the input features. 
	Specifically, 
	in the spatial GraphFormer, the input features are the  element-wise addition of the token and positional features within a frame.
	In the temporal GraphFormer, the input features are the multi-frame  intermediate features  output by the spatial GraphFormer.

	\begin{table*}[t]
		\renewcommand{\arraystretch}{0.9845}
		\centering
		\scalebox{0.9}{\tabcolsep0.135in
			\begin{tabular}{l|r|c|c|c|c} 
				\toprule
				Method  & \centering{Pub. \& Year~~~~\!} & Backbone & Base Detector  & mAP (\%) & Runtime (ms) \\
				\midrule 		  
				OGEM \cite{42}     & ICCV-2019~~~~\!      & ResNet-101    &R-FCN         & 79.3     &   112.0     \\
				TCENet \cite{17}   &AAAI-2020~~~~\!       & ResNet-101    &R-FCN         &80.3      &    125.0  \\
				MEGA \cite{37}  &CVPR-2020~~~~\!       & ResNet-101    &Faster R-CNN  &82.9    & 114.5 \\
				HVRNet \cite{49}   &ECCV-2020~~~~\!       & ResNet-101    & Faster R-CNN &  83.2     &  --    \\
				DSFNet \cite{50}  &ACM MM-2020~~~~\!     & ResNet-101    &Faster R-CNN  & 84.1  &    384.5          \\
				EBFA \cite{51}   &ACM MM-2020~~~~\!    & ResNet-101   &Faster R-CNN   & 84.8    & --   \\
				MINet \cite{53}        &TIP-2021~~~~\!       & ResNet-101    &Faster R-CNN  & 80.2   &  133.0 \\
				TransVOD \cite{11} &ACM MM-2021~~~~\!     & ResNet-101    &Deformable DETR          & 81.9   & $ > $341.1  \\
				MAMBA \cite{52}  &AAAI-2021~~~~\!       & ResNet-101    &Faster R-CNN  & 84.6   &  110.3 \\
				CSMN \cite{10}      &IJCV-2021~~~~\!    & ResNet-101    &Faster R-CNN        & 85.2   & 909.1\\
				EOVOD \cite{55}    &ECCV-2022~~~~\!       & ResNet-101    &Faster R-CNN  & 79.8   &  49.0 \\
				QueryProp \cite{54}     &AAAI-2022~~~~\!       & ResNet-101    &Sparse R-CNN  & 82.3  &  37.3 \\
				TSFA \cite{17}         &PR-2022~~~~\!       & ResNet-101    &Faster R-CNN  & 82.5   &  125.0 \\
				MSTF \cite{56}   &TCSVT-2022~~~~\!    & ResNet-101   &Faster R-CNN   & 83.3    & 105.7   \\
				CFANet \cite{57}   &TCSVT-2022~~~~\!    & ResNet-101   &Faster R-CNN   & 85.0    & 884.2   \\
				GMLCN \cite{58}          &TMM-2023~~~~\!        & ResNet-101    &Faster R-CNN  & 78.6  &  39.6 \\
				TransVOD++ \cite{8}       &TPAMI-2023~~~~\!      & ResNet-101    &Deformable DETR  &82.0  & -- \\
				ClipVID \cite{12}             &ICCV-2023~~~~\!      & ResNet-101    & DETR  &84.7  & 25.5 \\
				DGRNet \cite{9}            &TIP-2023~~~~\!      & ResNet-101    & Faster R-CNN  &85.0  & 91.7 \\
				CETR \cite{15}             &AAAI-2024~~~~\!      &   ResNet-101   & DAB-DETR  &79.6  & 42.9 \\
				CDANet \cite{21}             &TMM-2024~~~~\!      &   ResNet-101   & Faster R-CNN  &85.4  & 80.6 \\
				TGBFormer (ours)  &   --~~~~~~~~~~\!   & ResNet-101  &DETR&  \textbf{86.5}    &   \textbf{24.3}     \\
				\midrule 
				TransVOD Lite \cite{8}      &TPAMI-2023~~~~\!    & Swin-B   &    Deformable DETR    &   90.1 & 67.1\\
				TGBFormer (ours)  &   --~~~~~~~~~~\!   &  Swin-B  &DETR &  \textbf{90.3}   &    \textbf{49.7}     \\
				\bottomrule
		\end{tabular}}
			\caption{Performance comparison with 
		some state-of-the-art 
		video object detection methods
		on the ImageNet VID dataset.
		‘--’ denotes the corresponding results are not publicly available. 
		The best results are highlighted in the bold font.}
		\label{table 1}
	\end{table*}

	\subsection{Global-Local Feature Blender Module}
	\label{section 3.4}
	
	Given the fact that there exist indeterminate knowledge discrepancies between transformer global features and GraphFormer local features, 
	simply assembling (e.g., concatenating or adding)  these features is ineffective for significant performance improvement.
	To this end, we customize a global-local feature blender module,  with the  blender weights dynamically balanced depending on the input  transformer global features and GraphFormer local features.
	Compared to the simple ensemble of global and local features, 
	our elaborately-customized  global-local feature blender module tends to be more flexible and effective. 
	In particular,
	the calculation of the global-local feature blender module can be formulated by:
	\begin{equation}
		\boldsymbol{B} = \boldsymbol{\alpha}_{GF} \otimes \boldsymbol{G} + \boldsymbol{\alpha}_{LF} \otimes \boldsymbol{L},
		\label{equation 9}
	\end{equation}
	where $ \boldsymbol{B} \in \mathbb{R}^{D \times N\!\!\!~M}$ stands for the blended feature matrix of $N$ frames. $\boldsymbol{G} \in \mathbb{R}^{D \times N\!\!\!~M}$ and $\boldsymbol{L} \in \mathbb{R}^{D \times N\!\!\!~M}$ respectively represent the global aggregated feature matrix output by STTM and the local aggregated feature matrix output by STGM.
	The notation of $\otimes $ denotes the element-wise multiplication operator.
	$ \boldsymbol{\alpha}_{GF} \in \mathbb{R}^{D \times N\!\!\!~M} $ and $\boldsymbol{\alpha}_{LF} \in \mathbb{R}^{D \times N\!\!\!~M}$ represent the blender weights, which can be formulated 
	by:
	\begin{equation}
		\boldsymbol{\alpha}_{GF}, \boldsymbol{\alpha}_{LF}= \text{softmax}(\boldsymbol{W}_{\!\alpha} \!~\varPi \big[\boldsymbol{G}, \boldsymbol{L} \big]),
		\label{equation 10}
	\end{equation}
	where $\boldsymbol{W}_{\!\alpha} \in \mathbb{R}^{2D \times 2D}$ denotes the trainable parameters of one linear projection layer.
	$\varPi [\cdot,\cdot]$ is the concatenation operation between features.
	With the help of the elaborately-customized global-local feature blender module, 
	our TGBFormer is capable of boosting the collaborative representations from  transformer and GraphFormer to facilitate video object detection.

	\section{Experiments}
	
	\subsection{Dataset and Metric}
	\label{section 4.1}
	For fair and convincing comparisons, we conduct experiments on the ImageNet VID dataset \cite{32}.
	It  spans 30 object classes, and contains 3,862 training  and 555 validation videos.
	Following the same protocols in previous video object detection works \cite{15,21,14,17},
	we 
	evaluate our method on the validation set 
	and 
	adopt the mean average precision (mAP) at an Intersection-over-Union (IoU) threshold of 0.5 as the evaluation metric.

	\subsection{Implementation Details}
	\label{section 4.2}
	\textbf{Network Architecture}:
	We choose DETR \cite{26}  as the baseline detection network, and utilize
	ResNet-101 \cite{33}  as the backbone network for fair comparisons. 
	Following the common implementation protocols in  \cite{46,47,48,19,20}, 
	we  adjust the total stride of the last stage (i.e., Conv5) in ResNet-101 from 32 to 16.
	Besides,
	we adopt Swin transformer \cite{35}  as the backbone network for better   performance.
	\\
	\textbf{TGBFormer}:
	We employ a sinusoidal positional encoding as in DETR \cite{26} to yield positional features.
	The number of  object queries per frame is set as 80 when utilizing the ResNet-101 backbone and 64 when using the Swin transformer backbone. 
	The number of attention heads in the spatial and temporal transformer is set as 6 and the number of dynamic graph convolution layers in the spatial and temporal GraphFormer is set as 2.
	The real-value thresholds in our graph pruning mechanism  are set as $\varGamma=[0.1, 0.3, 1] $.
	The residual constant in Equation (\ref{equation 8}) is set as 0.5.
	\\
	\textbf{Training  and Testing Details}:
	Following the common practice in previous video object detection methods \cite{15,18,31}, we utilize both ImageNet DET and ImageNet VID datasets to train our TGBFormer, and resize each input frame to a shorter dimension of 600 pixels for fair comparisons.
	We train our TGBFormer  on 4 Nvidia Tesla A100 GPUs by using the AdamW optimizer. 
	The whole training procedure lasts for 150K iterations, with a learning rate of $10^{-4}$ for the first 110K iterations and $10^{-5}$ for the last 40K iterations.
	Besides,
	we also employ the same data augmentation  as the work of \cite{37} to make the training procedure more effective.
	At the testing phase,
	TGBFormer takes  consecutive $N$ (which is set as 25 by default) frames as inputs and outputs the detection results of all input frames at one stroke.

	\subsection{State-of-the-Art Comparison}
	\label{section 4.3}
	
	We compare our TGBFormer  with several existing video object detection methods, and summarize the results 
	in Table \ref{table 1}. 
	As illustrated in Table \ref{table 1}, we group the competing methods into two categories by their backbones. 
	Since most  video object detection methods are built on the backbone of the ResNet family, we first compare our TGBFormer with the other methods under the ResNet-101 backbone for fair comparisons.
	With the ResNet-101 backbone, our TGBFormer achieves 86.5\% mAP  at the runtime of 24.3 ms, 1.1\% absolute mAP improvement and more than $3.3\times$ faster over the leading method CDANet \cite{21}.
	This is because CDANet neglects the long-range temporal dependencies and employs a non-parallel frame-wise detection fashion, which yields  incomprehensive feature aggregation and slows down the running speed.
	Compared with  QueryProp \cite{54}, GMLCN \cite{58} and ClipVID \cite{12}, which are specifically optimized for efficient inference in real-world scenarios, our TGBFormer leads in both accuracy and runtime.
	This demonstrates that our TGBFormer is more friendly for real-world scenarios.
	The accuracy of our TGBFormer derives from customizing transformers and dynamic graph convolutional networks to perform more comprehensive feature aggregation, which considers both long-range and short-range spatial-temporal information and thus makes detection more robust to the issues like blur  or occlusion in videos.
	The efficiency of our TGBFormer comes from the parallel sequence-wise detection manner, which simultaneously detects objects on all input frames.
	
	Since
	our method is backbone agnostic,
	we also report the results of our method using a more advanced backbone Swin transformer Base (Swin-B), in which case
	our TGBFormer obtains 90.3\% mAP at runtime of 49.7 ms, which has a 0.2\% mAP improvement and more than $1.3\times$ faster over the competing method TransVOD Lite \cite{8}.

	\subsection{Ablation Study}
	\label{section 4.4}

\textbf{Effectiveness of each component in TGBFormer}:
To validate  the effectiveness of each component in our TGBFormer, we conduct  experiments to study how they contribute to the final  accuracy, and the results are presented in Table \ref{table 2}.

Method (a) denotes the baseline detector DETR \cite{26} using the ResNet-101 backbone, and it achieves 77.6\% overall mAP.

Method (b) integrates the spatial-temporal transformer module into (a), 
which brings a clear overall mAP improvement (from 77.6\% to 83.3\%). 

Method (c) incorporates a lite-version (i.e., without utilizing the graph pruning mechanism) of the spatial-temporal GraphFormer module into (a), which improves the overall mAP from 77.6\% to 82.5\%.
The reason is that the elaborately-customized spatial-temporal GraphFormer  module can effectively  aggregate short-range spatial and temporal information to facilitate video object detection and thus improve detection accuracy.

Method (d) adds the graph pruning mechanism into (c)  to attend the useful local relationships between nodes by  removing useless/weak connection edges. 
It leads to an increase of 0.6\% overall mAP against (c), 
indicating that the  graph pruning mechanism is effective  in our TGBFormer.

Method (e) adds  the spatial-temporal transformer module  into (d), and employs a vanilla concatenation operation to integrate the transformer global features and GraphFormer local features.
It achieves 85.4\% overall mAP, outperforming (d) by  2.3\% overall mAP,
which further demonstrates the effectiveness of the spatial-temporal transformer module.

Method (f) is the proposed TGBFormer, which introduces the global-local feature blender module into (e) to adaptively couple the  transformer global features and GraphFormer local features. 
It achieves 86.5\% overall mAP,  with a 1.1\%   improvement over (e), which validates the effectiveness of the global-local feature blender module.
Moreover, 
our TGBFormer significantly outperforms (b)-(e), indicating that all proposed components contribute to better   accuracy.
	\begin{table}
	\centering
	\scalebox{0.9}{\tabcolsep0.056in
		\begin{tabular}{c|c|c|c|c|c|p{1.1cm}<{\centering}}
			\toprule
			Method &   (a) &  (b)  & (c) &   (d)  & (e) &(f)  \\
			\midrule 	
			Transformer  &     & $\checkmark$  & & &  $\checkmark$ &   $\checkmark$  \\
			GraphFormer w/o GP &     &                              & $\checkmark$&        $\checkmark$          &{$\checkmark$}  & $\checkmark$  \\
			Global-Local Blender&      & & &&   &  $\checkmark$     \\
			Graph Pruning (GP)     &      & & &$\checkmark$ &  $\checkmark$ &  $\checkmark$ \\
			\midrule [0.5pt]
			mAP (\%) (overall)   & 77.6& 83.3   & 82.5 & 83.1& 85.4 &  $ \textbf{86.5}_{\uparrow \text{11.4}} $   \\ 
			\midrule [0.5pt]
			mAP (\%) (slow)     & 85.5 & 88.1    & 87.0& 87.9 &90.7  & $\textbf{91.8}_{\uparrow \text{7.7}} $  \\  
			mAP (\%) (medium)  & 75.6 & 81.7 & 80.6 & 81.2& 84.4& $ \textbf{85.6}_{\uparrow \text{13.2}} $  \\  
			mAP (\%) (fast)     & 56.7& 66.9&  65.4 & 66.5 & 70.6& $ \textbf{71.9}_{\uparrow \text{19.1}} $  \\  
			\bottomrule
	\end{tabular}}
	\caption{%
		Ablation studies on each component. 
		‘w/o' is the abbreviation for ‘without'.
		‘slow/medium/fast’ represents the objects with ‘slow/medium/fast' motions.
		The best results are boldfaced.
	}
\label {table 2}
\end{table}
%
%
\\		
\textbf{Analysis on the Number of Dynamic Graph Convolution  Layers in STGM}:
In the customized STGM, we leverage two successive dynamic graph convolution layers to implement the spatial and temporal GraphFormer, as shown in Figure \ref{figure 5} and Equation (\ref{equation 8}).
To comprehensively analyze the influence of the number of dynamic graph convolution layers on detection accuracy,  we conduct  some ablation studies in Table \ref{table 4}. 
As shown  in Table \ref{table 4},
the mAP improves consistently when the number of dynamic graph convolution layers $L_{DGC}$ increases, and it tends to  be optimal when the value of  $L_{DGC}$ is up to 2.
However, when further enlarging $L_{DGC}$ from 2 to 4, the mAP is decreased by 0.9\%  (from 86.5\% mAP to 85.6\% mAP). 
This is because
deep-layer graph topological  structures possibly bring several ambiguous or even mistaken connection edges, 
in which case conducting dynamic graph convolutions would  result in the aggregation of inter-class feature information or background information and thus inevitably decrease accuracy. 
\\
\textbf{Influence of the Number of Inference Frames in TGBFormer}:
At the inference/testing stage, we input consecutive $N$ inference frames into TGBFormer  to simultaneously predict the detection results of $N$ inference frames, and thus the number of inference frames $N$ is an important parameter that affects the accuracy and runtime of TGBFormer.
Table \ref{table 5} illustrates the influence of the parameter $N$.
The results demonstrate that
enlarging the value of $N$ leads to both accuracy improvements and runtime decreases, and the best accuracy is achieved when $N$ is set as 25.
Interestingly, we find that the accuracy slightly decreases when increasing the number of inference frames from 25 to 30, but the runtime is further improved thanks to the elaborately-customized sequence-wise detection paradigm.
The reason is that
feeding more inference frames into TGBFormer may hamper the learning of GraphFormer and prevent it from exploiting useful local information from long-range temporal contexts.
To strike the trade-off between accuracy and runtime, we select 25 as the default value of $N$.
\begin{table}[t] 
	\centering
	\scalebox{0.9}{\tabcolsep0.155in
			\begin{tabular} {c| c|c|c|c|c}
				\toprule 
				\# $\!L_{DGC} $ &  0&  1& $ \text{2}^{\divideontimes} $&3& 4\\ 
				\midrule [0.5pt]	
				mAP (\%) & 83.5&85.8 & \textbf{86.5} &86.1  &85.6 \\ 
				\bottomrule 
		\end{tabular}}
		\caption{%
		Analysis on the number of  dynamic graph convolution layers in STGM.
		The default parameter setting is represented by the symbol of $  \divideontimes $.
		The best result is boldfaced.
	}
		\label{table 4}
	\end{table}
	\begin{table}[t] 
		\renewcommand\arraystretch{1.0}
		\centering
		\scalebox{0.9}{\tabcolsep0.098in
			\begin{tabular} {c| c|c|c|c|c|c}
				\toprule 
				\# $\!N$ &  1&  10& 15&20& $ \text{25}^{\divideontimes} $&30\\ 
				\midrule [0.5pt]	
				mAP (\%) & 79.5&83.2 & 84.8&85.9&\textbf{86.5} & 86.3\\ 
				Runtime (ms) & 50.9& 40.5& 33.6&   27.9&   24.3 & \textbf{21.7}\\
				\bottomrule 
		\end{tabular}}
		\caption{%
	Influence of the number of  inference frames.
	$ \divideontimes $ denotes the default  setting.
	The best results are boldfaced.
}
		\label{table 5}
	\end{table}

	\section{Conclusion}
	\label{section 5}
	
	In this paper, we propose a novel Transformer-GraphFormer Blender Network (TGBFormer) for video object detection, which provides a new perspective on feature aggregation by combining the advantages of transformers and graph convolutional networks.
	Our TGBFormer includes three key technical improvements against existing video object detection methods.
	First, we customize a spatial-temporal transformer module to aggregate global contextual information by utilizing the long-range feature dependencies, which contributes to constituting the global representations of objects.
	Second, we design a spatial-temporal GraphFormer module to perform feature aggregation from the perspective of graph convolutions, which facilitates generating new local representations of objects that are complementary to the transformer outputs.
	Third, we develop a global-local feature blender module to adaptively couple transformer-style global features and GraphFormer-style local features, which is beneficial to generate comprehensive feature representations.
	We conduct extensive experiments on the public ImageNet VID dataset, and the results show that our TGBFormer 
	sets new state-of-the-art performance.
	
\section{Acknowledgments}
This work was supported in part by the Natural Science Foundation of Shandong Province under Grant 322024038; and in part by the Low-altitude Flight Intelligent Service Support Shandong Engineering 
Research Center under Grant 412024014.

	
	\bibliography{aaai25}

\begin{thebibliography}{46}
\providecommand{\natexlab}[1]{#1}

\bibitem[{An et~al.(2024)An, Park, Kim, Baek, Lee, and Kim}]{15}
An, S.; Park, S.; Kim, G.; Baek, J.; Lee, B.; and Kim, S. 2024.
\newblock Context Enhanced Transformer for Single Image Object Detection in
  Video Data.
\newblock In \emph{Proc. AAAI Conf. Artif. Intell. (AAAI)}, 682--690.

\bibitem[{Bertasius, Torresani, and Shi(2018)}]{20}
Bertasius, G.; Torresani, L.; and Shi, J. 2018.
\newblock Object Detection in Video with Spatiotemporal Sampling Networks.
\newblock In \emph{Proc. Eur. Conf. Comput. Vis. (ECCV)}, 331--346.

\bibitem[{Carion et~al.(2020)Carion, Massa, Synnaeve, Usunier, Kirillov, and
  Zagoruyko}]{26}
Carion, N.; Massa, F.; Synnaeve, G.; Usunier, N.; Kirillov, A.; and Zagoruyko,
  S. 2020.
\newblock End-to-end object detection with transformers.
\newblock In \emph{Proc. Eur. Conf. Comput. Vis. (ECCV)}, 213--229.

\bibitem[{Chen et~al.(2023)Chen, Pan, Lu, Fan, and Li}]{22}
Chen, X.; Pan, J.; Lu, J.; Fan, Z.; and Li, H. 2023.
\newblock Hybrid cnn-transformer feature fusion for single image deraining.
\newblock In \emph{Proc. AAAI Conf. Artif. Intell. (AAAI)}, 378--386.

\bibitem[{Chen et~al.(2020)Chen, Cao, Hu, and Wang}]{37}
Chen, Y.; Cao, Y.; Hu, H.; and Wang, L. 2020.
\newblock Memory enhanced global-local aggregation for video object detection.
\newblock In \emph{Proc. IEEE Conf. Comput. Vis. Pattern Recognit. (CVPR)},
  10337--10346.

\bibitem[{Chitta et~al.(2023)Chitta, Prakash, Jaeger, Yu, Renz, and Geiger}]{2}
Chitta, K.; Prakash, A.; Jaeger, B.; Yu, Z.; Renz, K.; and Geiger, A. 2023.
\newblock TransFuser: Imitation With Transformer-Based Sensor Fusion for
  Autonomous Driving.
\newblock \emph{IEEE Trans. Pattern Anal. Mach. Intell.}, 45(11): 12878--12895.

\bibitem[{Cui(2023)}]{14}
Cui, Y. 2023.
\newblock Feature Aggregated Queries for Transformer-Based Video Object
  Detectors.
\newblock In \emph{Proc. IEEE Conf. Comput. Vis. Pattern Recognit. (CVPR)},
  6365--6376.

\bibitem[{Cui et~al.(2021)Cui, Yan, Cao, and Liu}]{48}
Cui, Y.; Yan, L.; Cao, Z.; and Liu, D. 2021.
\newblock TF-Blender: Temporal Feature Blender for Video Object Detection.
\newblock In \emph{Proc. IEEE Int. Conf. Comput. Vis. (ICCV)}, 8138--8147.

\bibitem[{Deng, Chen, and Wu(2023)}]{12}
Deng, C.; Chen, D.; and Wu, Q. 2023.
\newblock Identity-Consistent Aggregation for Video Object Detection.
\newblock In \emph{Proc. IEEE Int. Conf. Comput. Vis. (ICCV)}, 13434--13444.

\bibitem[{Deng et~al.(2019)Deng, Hua, Song, Zhang, Xue, Ma, Robertson, and
  Guan}]{42}
Deng, H.; Hua, Y.; Song, T.; Zhang, Z.; Xue, Z.; Ma, R.; Robertson, N.; and
  Guan, H. 2019.
\newblock Object guided external memory network for video object detection.
\newblock In \emph{Proc. IEEE Int. Conf. Comput. Vis. (ICCV)}, 6678--6687.

\bibitem[{Deng et~al.(2020)Deng, Pan, Yao, Zhou, Li, and Mei}]{47}
Deng, J.; Pan, Y.; Yao, T.; Zhou, W.; Li, H.; and Mei, T. 2020.
\newblock Single shot video object detector.
\newblock \emph{IEEE Trans. Multimedia}, 23: 846--858.

\bibitem[{Deng et~al.(2021)Deng, Pan, Yao, Zhou, Li, and Mei}]{53}
Deng, J.; Pan, Y.; Yao, T.; Zhou, W.; Li, H.; and Mei, T. 2021.
\newblock MINet: Meta-Learning Instance Identifiers for Video Object Detection.
\newblock \emph{IEEE Trans. Image Process.}, 30: 6879--6891.

\bibitem[{Han et~al.(2020{\natexlab{a}})Han, Wang, Yin, Wang, and Li}]{51}
Han, L.; Wang, P.; Yin, Z.; Wang, F.; and Li, H. 2020{\natexlab{a}}.
\newblock Exploiting better feature aggregation for video object detection.
\newblock In \emph{Proc. ACM Int. Conf. Multim. (ACM MM)}, 1469--1477.

\bibitem[{Han et~al.(2021)Han, Wang, Yin, Wang, and Li}]{10}
Han, L.; Wang, P.; Yin, Z.; Wang, F.; and Li, H. 2021.
\newblock Context and structure mining network for video object detection.
\newblock \emph{Int. J. Comput. Vis.}, 129: 2927--2946.

\bibitem[{Han et~al.(2022)Han, Wang, Yin, Wang, and Li}]{57}
Han, L.; Wang, P.; Yin, Z.; Wang, F.; and Li, H. 2022.
\newblock Class-aware Feature Aggregation Network for Video Object Detection.
\newblock \emph{IEEE Trans. Circuits Syst. Video Technol.}, 32(12): 8165--8178.

\bibitem[{Han and Yin(2023)}]{58}
Han, L.; and Yin, Z. 2023.
\newblock Global Memory and Local Continuity for Video Object Detection.
\newblock \emph{IEEE Trans. Multimedia}, 5: 3681--3693.

\bibitem[{Han et~al.(2020{\natexlab{b}})Han, Wang, Chang, and Qiao}]{49}
Han, M.; Wang, Y.; Chang, X.; and Qiao, Y. 2020{\natexlab{b}}.
\newblock Mining inter-video proposal relations for video object detection.
\newblock In \emph{Proc. Eur. Conf. Comput. Vis. (ECCV)}, 431--446.

\bibitem[{He et~al.(2022{\natexlab{a}})He, Gao, Jia, Zhao, and Huang}]{54}
He, F.; Gao, N.; Jia, J.; Zhao, X.; and Huang, K. 2022{\natexlab{a}}.
\newblock QueryProp: Object Query Propagation for High-Performance Video Object
  Detection.
\newblock In \emph{Proc. AAAI Conf. Artif. Intell. (AAAI)}, 2620--2627.

\bibitem[{He et~al.(2022{\natexlab{b}})He, Li, Zhao, and Huang}]{17}
He, F.; Li, Q.; Zhao, X.; and Huang, K. 2022{\natexlab{b}}.
\newblock Temporal-adaptive sparse feature aggregation for video object
  detection.
\newblock \emph{Pattern Recognit.}, 127: 108587.

\bibitem[{He et~al.(2016)He, Zhang, Ren, and Sun}]{33}
He, K.; Zhang, X.; Ren, S.; and Sun, J. 2016.
\newblock Deep residual learning for image recognition.
\newblock In \emph{Proc. IEEE Conf. Comput. Vis. Pattern Recognit. (CVPR)},
  770--778.

\bibitem[{He et~al.(2021)He, Zhou, Li, Niu, Cheng, Li, Liu, Tong, Ma, and
  Zhang}]{11}
He, L.; Zhou, Q.; Li, X.; Niu, L.; Cheng, G.; Li, X.; Liu, W.; Tong, Y.; Ma,
  L.; and Zhang, L. 2021.
\newblock End-to-End Video Object Detection with Spatial-Temporal Transformers.
\newblock In \emph{Proc. ACM Int. Conf. Multimedia (ACM MM)}, 1507--1516.

\bibitem[{He et~al.(2024)He, Wu, Huang, Hu, Wang, Sangiovanni-Vincentelli, and
  Lv}]{1}
He, X.; Wu, J.; Huang, Z.; Hu, Z.; Wang, J.; Sangiovanni-Vincentelli, A.; and
  Lv, C. 2024.
\newblock Fear-Neuro-Inspired Reinforcement Learning for Safe Autonomous
  Driving.
\newblock \emph{IEEE Trans. Pattern Anal. Mach. Intell.}, 46(1): 267--279.

\bibitem[{Jiang et~al.(2019)Jiang, Gao, Guo, Zhang, Xiang, and Pan}]{16}
Jiang, Z.; Gao, P.; Guo, C.; Zhang, Q.; Xiang, S.; and Pan, C. 2019.
\newblock Video object detection with locally-weighted deformable neighbors.
\newblock In \emph{Proc. AAAI Conf. Artif. Intell. (AAAI)}, 8529--8536.

\bibitem[{Jiang et~al.(2020)Jiang, Liu, Yang, Liu, Zhang, Xiang, and Pan}]{46}
Jiang, Z.; Liu, Y.; Yang, C.; Liu, J.; Zhang, Q.; Xiang, S.; and Pan, C. 2020.
\newblock Learning where to focus for efficient video object detection.
\newblock In \emph{Proc. Eur. Conf. Comput. Vis. (ECCV)}, 18--34.

\bibitem[{Kipf and Welling(2017)}]{25}
Kipf, T.~N.; and Welling, M. 2017.
\newblock Semi-supervised classification with graph convolutional networks.
\newblock In \emph{Proc. Int. Conf. Learn. Representations (ICLR)}, 565--578.

\bibitem[{Li et~al.(2019)Li, Zhang, Chen, and Huang}]{3}
Li, D.; Zhang, Z.; Chen, X.; and Huang, K. 2019.
\newblock A Richly Annotated Pedestrian Dataset for Person Retrieval in Real
  Surveillance Scenarios.
\newblock \emph{IEEE Trans. Image Process.}, 28(4): 1575--1590.

\bibitem[{Lin et~al.(2020)Lin, Chen, Zhang, Liang, Li, Shan, and Wang}]{50}
Lin, L.; Chen, H.; Zhang, H.; Liang, J.; Li, Y.; Shan, Y.; and Wang, H. 2020.
\newblock Dual semantic fusion network for video object detection.
\newblock In \emph{Proc. ACM Int. Conf. Multimedia (ACM MM)}, 1855--1863.

\bibitem[{Liu, Wu, and Jia(2016)}]{5}
Liu, C.; Wu, X.; and Jia, Y. 2016.
\newblock A hierarchical video description for complex activity understanding.
\newblock \emph{Int. J. Comput. Vis.}, 118: 240--255.

\bibitem[{Liu et~al.(2020)Liu, Shahroudy, Perez, Wang, Duan, and Kot}]{6}
Liu, J.; Shahroudy, A.; Perez, M.; Wang, G.; Duan, L.-Y.; and Kot, A.~C. 2020.
\newblock NTU RGB+D 120: A Large-Scale Benchmark for 3D Human Activity
  Understanding.
\newblock \emph{IEEE Trans. Pattern Anal. Mach. Intell.}, 42(10): 2684--2701.

\bibitem[{Liu et~al.(2021)Liu, Lin, Cao, Hu, Wei, Zhang, Lin, and Guo}]{35}
Liu, Z.; Lin, Y.; Cao, Y.; Hu, H.; Wei, Y.; Zhang, Z.; Lin, S.; and Guo, B.
  2021.
\newblock Swin Transformer: Hierarchical Vision Transformer Using Shifted
  Windows.
\newblock In \emph{Proc. IEEE Int. Conf. Comput. Vis. (ICCV)}, 10012--10022.

\bibitem[{Peng et~al.(2023)Peng, Guo, Huang, Wang, Xie, Jiao, Tian, and
  Ye}]{23}
Peng, Z.; Guo, Z.; Huang, W.; Wang, Y.; Xie, L.; Jiao, J.; Tian, Q.; and Ye, Q.
  2023.
\newblock Conformer: Local Features Coupling Global Representations for
  Recognition and Detection.
\newblock \emph{IEEE Trans. Pattern Anal. Mach. Intell.}, 45(8): 9454--9468.

\bibitem[{Qi et~al.(2023{\natexlab{a}})Qi, Hou, Lu, Yan, and Wang}]{9}
Qi, Q.; Hou, T.; Lu, Y.; Yan, Y.; and Wang, H. 2023{\natexlab{a}}.
\newblock DGRNet: A Dual-Level Graph Relation Network for Video Object
  Detection.
\newblock \emph{IEEE Trans. Image Process.}, 32: 4128--4141.

\bibitem[{Qi et~al.(2023{\natexlab{b}})Qi, Hou, Yan, Lu, and Wang}]{18}
Qi, Q.; Hou, T.; Yan, Y.; Lu, Y.; and Wang, H. 2023{\natexlab{b}}.
\newblock TCNet: A Novel Triple-Cooperative Network for Video Object Detection.
\newblock \emph{IEEE Trans. Circuits Syst. Video Technol.}, 33(8): 3649--3662.

\bibitem[{Qi, Yan, and Wang(2024)}]{21}
Qi, Q.; Yan, Y.; and Wang, H. 2024.
\newblock Class-Aware Dual-Supervised Aggregation Network for Video Object
  Detection.
\newblock \emph{IEEE Trans. Multimedia}, 26: 2109--2123.

\bibitem[{Qianyu et~al.(2023)Qianyu, Li, He, Yang, Cheng, Tong, Ma, and
  Tao}]{8}
Qianyu, Z.; Li, X.; He, L.; Yang, Y.; Cheng, G.; Tong, Y.; Ma, L.; and Tao, D.
  2023.
\newblock TransVOD: End-to-End Video Object Detection With Spatial-Temporal
  Transformers.
\newblock \emph{IEEE Trans. Pattern Anal. Mach. Intell.}, 45(6): 7853--7869.

\bibitem[{Russakovsky et~al.(2015)Russakovsky, Deng, Su, Krause, Satheesh, Ma,
  Huang, Karpathy, Khosla, Bernstein et~al.}]{32}
Russakovsky, O.; Deng, J.; Su, H.; Krause, J.; Satheesh, S.; Ma, S.; Huang, Z.;
  Karpathy, A.; Khosla, A.; Bernstein, M.; et~al. 2015.
\newblock Imagenet large scale visual recognition challenge.
\newblock \emph{Int. J. Comput. Vis.}, 115: 211--252.

\bibitem[{Sun et~al.(2021)Sun, Hua, Hu, and Robertson}]{52}
Sun, G.; Hua, Y.; Hu, G.; and Robertson, N. 2021.
\newblock MAMBA: Multi-level Aggregation via Memory Bank for Video Object
  Detection.
\newblock In \emph{Proc. AAAI Conf. Artif. Intell. (AAAI)}, 2620--2627.

\bibitem[{Sun et~al.(2022)Sun, Hua, Hu, and Robertson}]{55}
Sun, G.; Hua, Y.; Hu, G.; and Robertson, N. 2022.
\newblock Efficient one-stage video object detection by exploiting temporal
  consistency.
\newblock In \emph{Proc. Eur. Conf. Comput. Vis. (ECCV)}, 1--16.

\bibitem[{Wang et~al.(2022)Wang, Tang, Liu, Guan, Xie, and Song}]{13}
Wang, H.; Tang, J.; Liu, X.; Guan, S.; Xie, R.; and Song, L. 2022.
\newblock Ptseformer: Progressive temporal-spatial enhanced transformer towards
  video object detection.
\newblock In \emph{Proc. Eur. Conf. Comput. Vis. (ECCV)}, 732--747.

\bibitem[{Wu et~al.(2019)Wu, Chen, Wang, and Zhang}]{31}
Wu, H.; Chen, Y.; Wang, N.; and Zhang, Z. 2019.
\newblock Sequence level semantics aggregation for video object detection.
\newblock In \emph{Proc. IEEE Int. Conf. Comput. Vis. (ICCV)}, 9217--9225.

\bibitem[{Xu et~al.(2022)Xu, Zhang, Wang, Tian, and Liu}]{56}
Xu, C.; Zhang, J.; Wang, M.; Tian, G.; and Liu, Y. 2022.
\newblock Multi-level Spatial-temporal Feature Aggregation for Video Object
  Detection.
\newblock \emph{IEEE Trans. Circuits Syst. Video Technol.}, 32(11): 7809--7820.

\bibitem[{Yoo et~al.(2023)Yoo, Kim, Lee, Kim, Lee, and Kim}]{24}
Yoo, J.; Kim, T.; Lee, S.; Kim, S.~H.; Lee, H.; and Kim, T.~H. 2023.
\newblock Enriched CNN-Transformer Feature Aggregation Networks for
  Super-Resolution.
\newblock In \emph{Proc. IEEE Win. Conf. App. Comput. Vis. (WACV)}, 4956--4965.

\bibitem[{Yun et~al.(2019)Yun, Jeong, Kim, Kang, and Kim}]{30}
Yun, S.; Jeong, M.; Kim, R.; Kang, J.; and Kim, H.~J. 2019.
\newblock Graph Transformer Networks.
\newblock In \emph{Proc. Adv. Neural Inf. Process. Syst. (NeurIPS)},
  11983--11993.

\bibitem[{Zhang et~al.(2022)Zhang, Jia, Hu, and Tan}]{4}
Zhang, J.; Jia, X.; Hu, J.; and Tan, K. 2022.
\newblock Moving Vehicle Detection for Remote Sensing Video Surveillance With
  Nonstationary Satellite Platform.
\newblock \emph{IEEE Trans. Pattern Anal. Mach. Intell.}, 44(9): 5185--5198.

\bibitem[{Zhu et~al.(2021)Zhu, Su, Lu, Li, Wang, and Dai}]{64}
Zhu, X.; Su, W.; Lu, L.; Li, B.; Wang, X.; and Dai, J. 2021.
\newblock Deformable DETR: Deformable Transformers for End-to-End Object
  Detection.
\newblock In \emph{Proc. Int. Conf. Learn. Represent. (ICLR)}, 1--16.

\bibitem[{Zhu et~al.(2017)Zhu, Wang, Dai, Yuan, and Wei}]{19}
Zhu, X.; Wang, Y.; Dai, J.; Yuan, L.; and Wei, Y. 2017.
\newblock Flow-guided feature aggregation for video object detection.
\newblock In \emph{Proc. IEEE Int. Conf. Comput. Vis. (ICCV)}, 408--417.

\end{thebibliography}

\end{document}